\documentclass[runningheads]{llncs}
\usepackage{graphicx}
\usepackage{caption}
\usepackage{subcaption}
\usepackage{booktabs}
\usepackage{caption} 
\begin{document}
\title{User Adaptive Language Learning Chatbots with a Curriculum}
%
%
\author{Kun Qian\inst{1} \and
Ryan Shea\inst{1}\and
Yu Li\inst{1}\and
Luke Kutszik Fryer\inst{2} \and
Zhou Yu\inst{1,3}}
\authorrunning{K. Qian et al.}
\institute{Columbia University, New York, USA \\
\email{\{kq2157, rs4235, yl5016, zy2461\}@columbia.edu} \and
The University of Hong Kong, Hong Kong, China \\
\email{fryer@hku.hk} \and
Articulate.AI Inc, New York, USA\\
}
\maketitle              
\begin{abstract}
Along with the development of systems for natural language understanding and generation, dialog systems have been widely adopted for language learning and practicing.
Many current educational dialog systems perform chitchat, where the generated content and vocabulary are not constrained.
However, for learners in a school setting, practice through dialog is more effective if it aligns with students' curriculum and focuses on textbook vocabulary.
Therefore, we adapt lexically constrained decoding to a dialog system, which urges the dialog system to include curriculum-aligned words and phrases in its generated utterances.
We adopt a generative dialog system, BlenderBot3, as our backbone model
and evaluate our curriculum-based dialog system with middle school students learning English as their second language.
The constrained words and phrases are derived from their textbooks, suggested by their English teachers.
The evaluation result demonstrates that the dialog system with curriculum infusion improves students' understanding of target words and increases their interest in practicing English.
\end{abstract}

\section{Introduction}

Finding a consistent speaking partner can be challenging for language learners. 
However, chatbots offer a solution by providing an interactive environment for practice.
Traditional chatbots that rely on pre-written scripts often produce utterances that are limited and unresponsive~\cite{Ruan2021EnglishBotAA,Kuhail2022InteractingWE}.
Recent advancements in large pre-trained language models have led to the development of more adaptable conversational AI that can respond more naturally to user input~\cite{gpt2,t5,gpt3}.
However, these systems are primarily focused on casual conversation and lack a structured curriculum.
On the other hand, non-native speakers usually learn new languages using textbooks, and it is more helpful if the chatbot generates utterances based on a curriculum.


Our proposed solution is to create a user-adaptive language learning chatbot that incorporates an English language learning curriculum. 
The chatbot is designed to assist learners in practicing different language skills, such as grammar and vocabulary, specified by the curriculum. 
We use the pre-trained language model Blenderbot3~\cite{bb3} as a foundation and propose a multi-turns grid beam search to include curriculum-specific words and phrases during the decoding stage. 
To evaluate the effectiveness of our chatbot, we conduct a study with 155 8th-grade students using a curriculum developed based on their textbook and consultation with their teachers.
Our experimental results show that:
\begin{enumerate}
    \item Our curriculum-based chatbot increases the frequency of specified words or phrases used throughout dialogs on both the system side and the user side.
    \item The user-adaptive curriculum improves users' engagement and interest in our chatbot. 
    \item In general, our chatbot helps students correctly understand and use specified words or phrases.
\end{enumerate}
We will release the source code of our chatbot toolkit in the future so that teachers and students can easily design their own curriculum for learning and practicing.

\section{Related Work}
\subsection{Constrained Decoding}
Many natural language processing tasks require including or excluding specified words or phrases in output sequences, such as machine translation~\cite{grid}, summarization~\cite{summarization} and image captioning~\cite{caption}. Traditional methods of lexically constrained decoding mainly involved post-editing~\cite{pe_specia} and interactive prediction~\cite{ip_foster,ip_barrachina}. 
However, with the widely adopted beam search decoding method~\cite{beam}, Hokamp and Liu introduced grid beam search~\cite{grid}, which was the first to consider constraints throughout the decoding process. 
In addition to beam search, Anderson et al.~\cite{caption} proposed using a finite state machine to trace the satisfactory state of constraints.
Although grid beam search is effective at enforcing lexical constraints, it consumes more time and computation than regular decoding. To improve efficiency, techniques such as dynamic beam allocation~\cite{dba} and vectorized dynamic beam allocation~\cite{vec_dba} were introduced. 
Lu et al.~\cite{neuralogic,a*} also introduced the concept of neuralogic, which allows for the inclusion of more complex words or phrases in the generation process. 
With the development of masked language models, non-autoregressive decoding~\cite{Gu2017NonAutoregressiveNM} was proposed as a way to speed up the decoding process. Instead of generating sequences from left to right, non-autoregressive decoding generates tokens in parallel. 
Inspired by the Levenshtein Transformer~\cite{Gu2019LevenshteinT}, which iteratively refines generated sequences, Susanto et al.~\cite{c_lev} proposed inserting constraint tokens during refinement.
Xu and Carpuat~\cite{editor} further improved performance by reordering constraint words. 

\subsection{Generative Language Model}
A generative language model is a type of machine-learning model that is trained to generate text. This can include natural language text, such as writing, speech, or code. 
Early NLP models were based on rule-based systems, where a set of predefined rules were used to analyze and generate text~\cite{chomsky2009syntactic,fass1991computational,riloff2000rule}. These systems were limited in their ability to handle variations in language and were not able to learn from data.
Statistical models, such as n-gram models, were then developed to overcome the limitations of rule-based systems. These models used large amounts of text data to estimate the probability of a word or sequence of words occurring.
With the development of deep learning, neural network-based models were proposed for NLP tasks such as language modeling, machine translation, and text summarization. These models, such as LSTM~\cite{hochreiter1997long}
, were able to learn complex patterns in language and achieve state-of-the-art results on a wide range of NLP tasks.
Recently, transformer-based models~\cite{vaswani2017attention} such as BERT~\cite{bert} and GPT-2~\cite{gpt2} have been developed, marking a significant step forward in generative language models. This architecture enables models to process entire sentences or paragraphs at once and better capture long-term dependencies. Furthermore, additional works demonstrate that pre-training larger models with more raw data significantly improves performance on downstream tasks~\cite{opt,gpt3}. In this paper, we choose BlenderBot3~\cite{bb3} as our base model since it is pre-trained on large amounts of raw data and finetuned on dialog-specific data.

\subsection{Educational Chatbot}
Applying chatbots in educational field is increasing in popularity as they can provide instant feedback and personalized guidance to students without expensive costs. Another advantage of chatbots in education is that they are available 24/7, providing students with access to learning resources and support outside of regular class hours. This is particularly important in classes with a large number of students, where individual support from educators during classes can be difficult~\cite{Winkler2018UnleashingTP,Kuhail2022InteractingWE}. 
The early applications of chatbots in education were focused on providing students with access to information and resources, such as answering questions about the curriculum or providing definitions of terms~\cite{sinha2020educational}. These chatbots were typically based on rule-based systems and were not able to understand natural language input or provide personalized feedback.
With the advancement of machine learning and natural language processing techniques, chatbots became more sophisticated and were able to understand and respond to more complex student input. This led to the development of more advanced chatbots that were able to provide personalized feedback and guidance to students~\cite{benotti2017tool,clarizia2018chatbot,ranoliya2017chatbot}.
In recent years, there has been a growing interest in using chatbots in online and distance learning to provide students with individualized support and to improve engagement. Chatbots are also increasingly used in language learning, to provide students with personalized feedback and guidance, and to adapt to students' proficiency level and learning style~\cite{durall2020co,Zhao2020APIHelperHJ}.


\section{Methodology}
\subsection{Multi-Turns Grid Beam Search}
\label{sec:grid}
Previous works applying grid beam search, such as machine translation or image captioning, only need to generate once to complete tasks. However, for dialog systems, we do not require systems to generate all constraint words in one single turn and continue repeating those words throughout the dialog, as this would result in an unnatural conversation. Instead, we expect our chatbot to mention constraint words over the course of several dialogue turns, when suitable. Therefore, we propose a multi-turns grid beam search method, to adapt constrained decoding to a dialog setting.
\begin{figure}[h]
\vspace{-0.3cm}
\centering
\includegraphics[width=0.9\textwidth]{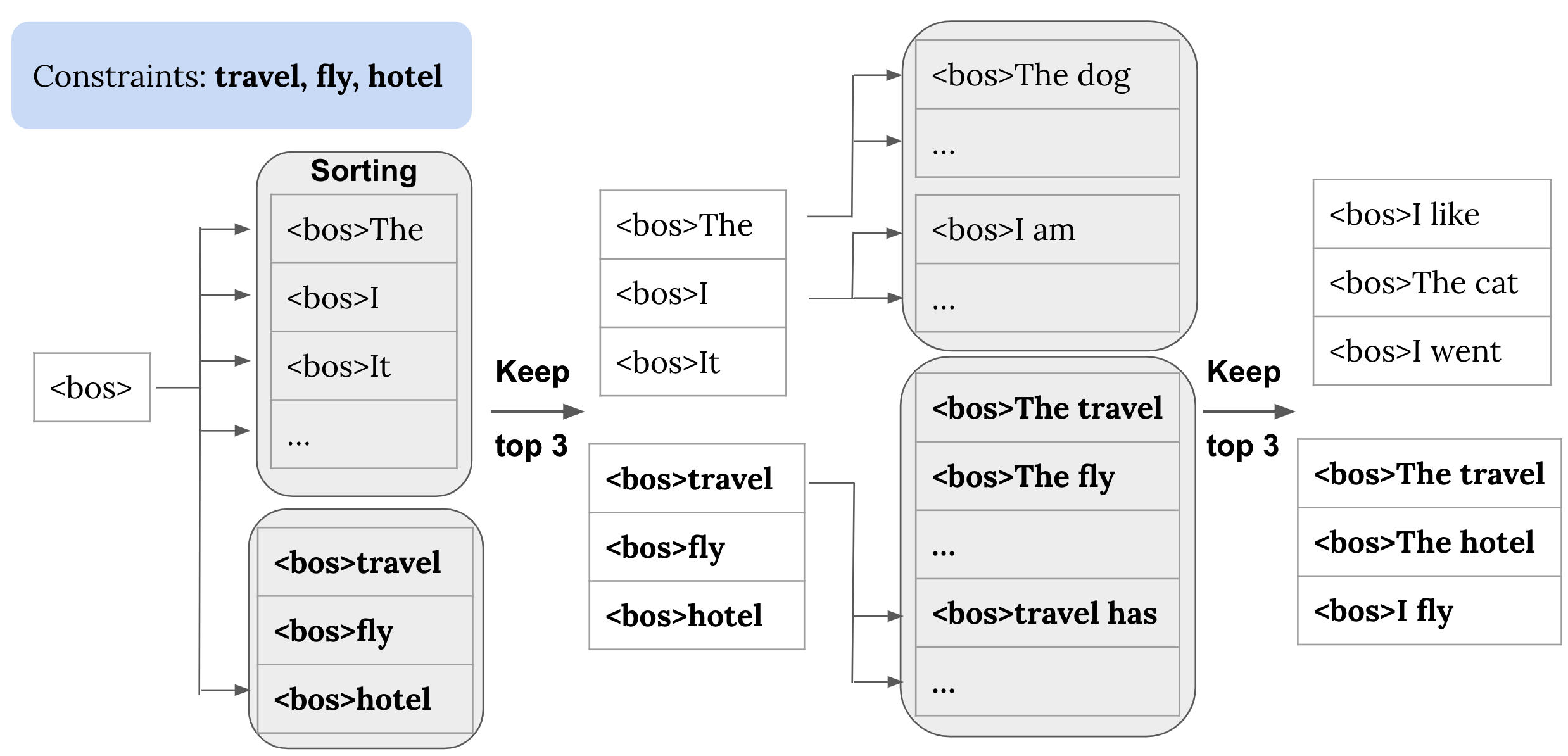}
\caption{First two decoding steps of multi-turns grid beam search. In addition to the normal beam (text in normal font), grid beam search adds an extra beam (text bolded) to store candidates that satisfy constraints. The constraint words are ``\textit{travel}'', ``\textit{fly}'', and ``\textit{hotel}'' in this case.} 
\label{fig: grid}
\vspace{-0.5cm}
\end{figure}

\noindent\textbf{Simplifed Constraint Beam Box.} Fig.~\ref{fig: grid} illustrates an example decoding step of multi-turns grid beam search. The upper beam boxes, where text is in normal font, are unsatisfied candidates. These candidates do not contain any constraint words and are selected following the normal beam search mechanism. The lower beam boxes, containing bolded text, are satisfied candidates including constraint words. They are generated by appending constraint words to the left candidates from the previous step. For example, at the first step, we append constraint words to the start token ``$<bos>$'' to get satisfied candidates ``$<bos>$ \textit{travel}'', ``$<bos>$ \textit{fly}'' and ``$<bos>$ \textit{hotel}''. Normally, we add extra beams based on the number of satisfied constraints, meaning that candidates in the same beam have the same number of constraint words. However, since one constraint word each turn is enough in our dialog setting, we fix the beam number as two (one for unsatisfied candidates and one for candidates containing one constraint word) and only append constraint words to the upper beam box. This modification helps speed up decoding.

\noindent\textbf{Dynamic Constraint Threshold.}
After appending constraint words, we sort two beam boxes separately based on candidates' accumulated probabilities and leave the top k candidates where k is the beam size. When reaching an ending token or the maximum decoding length, we compare the candidates from the two beam boxes. If the probability of satisfied candidates is higher than that of unsatisfied candidates minus a certain threshold value:
$$P(\textrm{satisfied cand})>P(\textrm{unsatisfied cand})-T$$
we accept the satisfied candidate. Otherwise, the unsatisfied candidate is far more natural and matches better with the dialog context. In this case, we generate the unsatisfied candidate.
In order to force our model to generate constraint words before a dialog ends, we propose a dynamic constraint threshold, where the threshold value increases by dialog turn number:
$$T=T_0\cdot 2\sigma(a\cdot t)$$
where $T_0$ is the initial threshold value,  $t$ is the turn number, $a$ is a scalar, and $$\sigma(x)=\frac{1}{1+e^{-x}}$$

\begin{figure}[h]
\vspace{-0.7cm}
\centering
\includegraphics[width=0.7\textwidth]{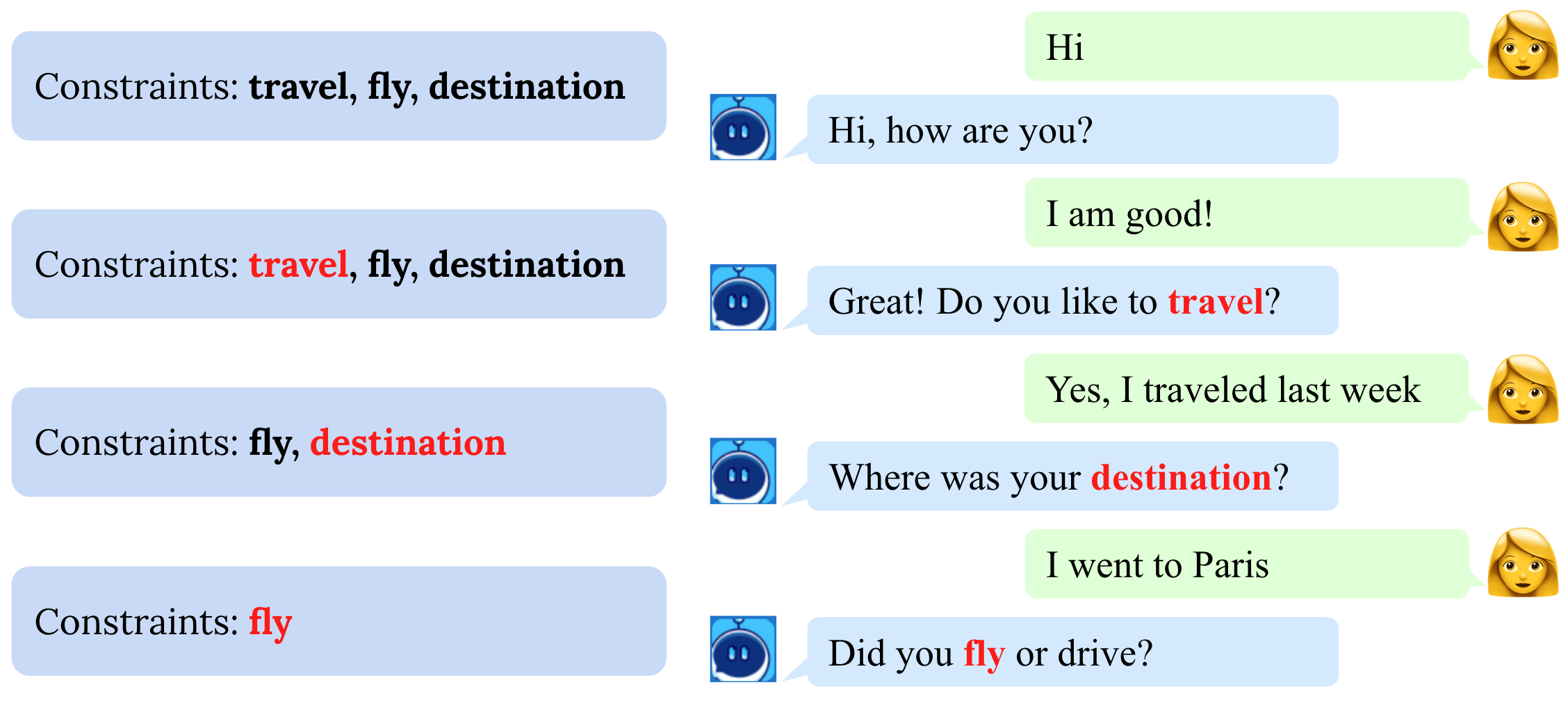}
\caption{The constraint list is updated through the conversation.} 
\label{fig: update}
\end{figure}

\noindent\textbf{Dynamic Constraint List.}
 Since mentioning constraint words once per dialog is enough, we update our constraint word list dynamically throughout the  dialog. As shown in Fig.~\ref{fig: update}, we check if any constraint word is used at the end of each turn and remove those words from the constraint list. These words are still likely to be generated due to the attention on dialog history, but we do not force the chatbot to generate them in future turns.

\vspace{-0.3cm}
\begin{figure}[]
     \centering
     \begin{subfigure}[b]{0.45\textwidth}
         \centering
         \includegraphics[width=\textwidth]{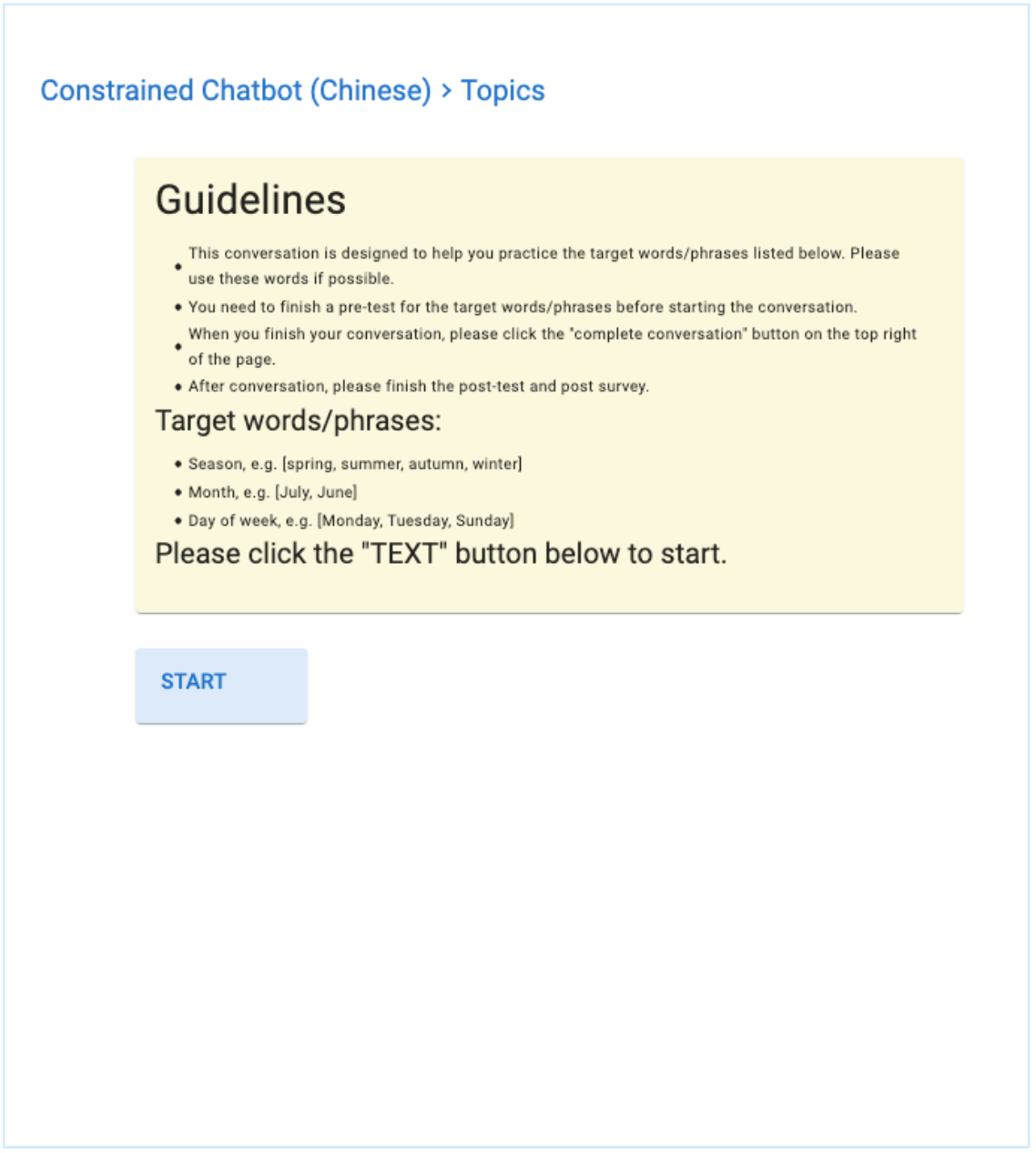}
         \caption{Instruction}
         \label{fig: instruction}
     \end{subfigure}
     \hfill
     \begin{subfigure}[b]{0.45\textwidth}
         \centering
         \includegraphics[width=\textwidth]{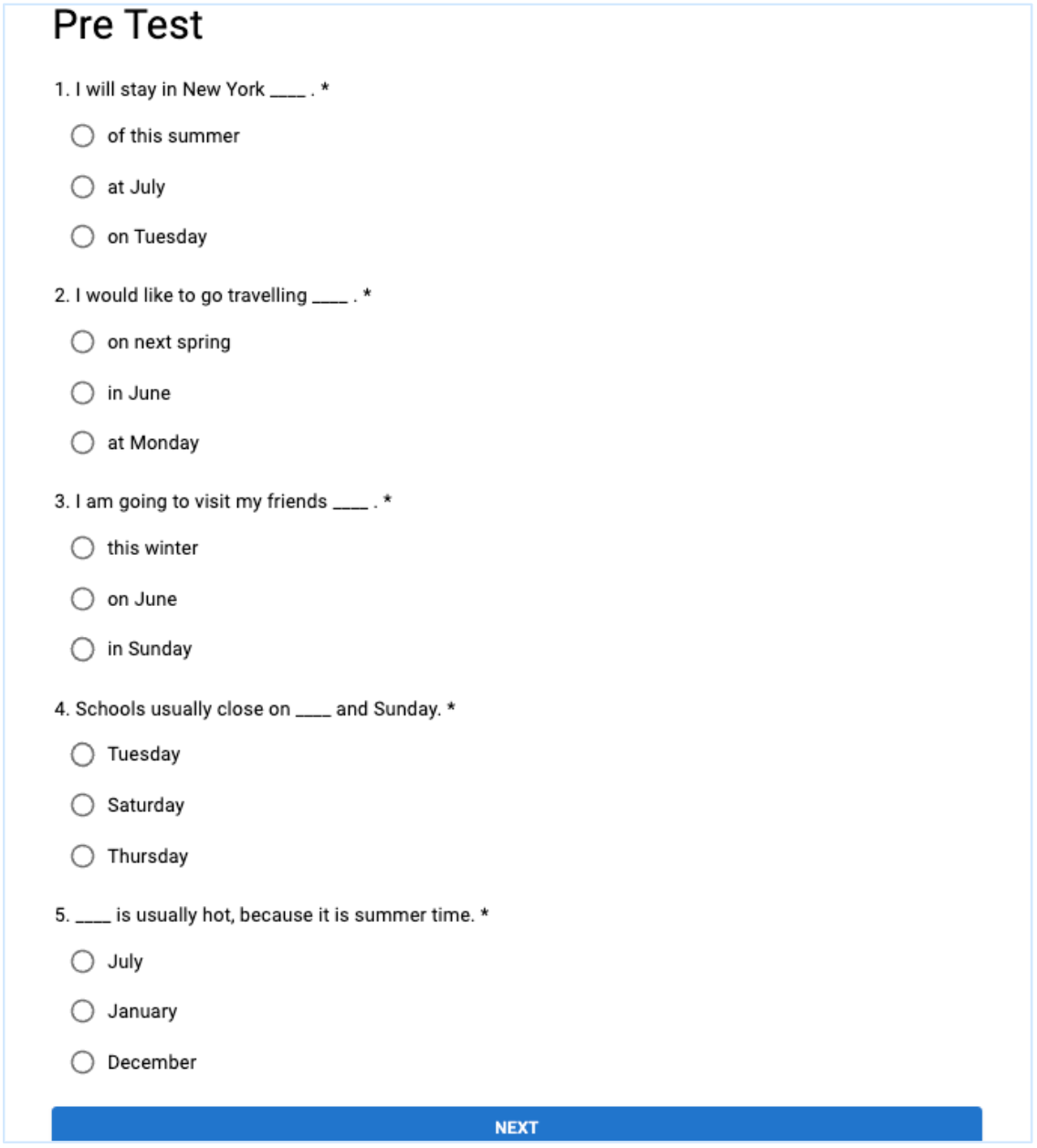}
         \caption{Pre-Test}
         \label{fig: pre-test}
     \end{subfigure}
     \vfill
     \begin{subfigure}[b]{0.45\textwidth}
         \centering
         \includegraphics[width=\textwidth]{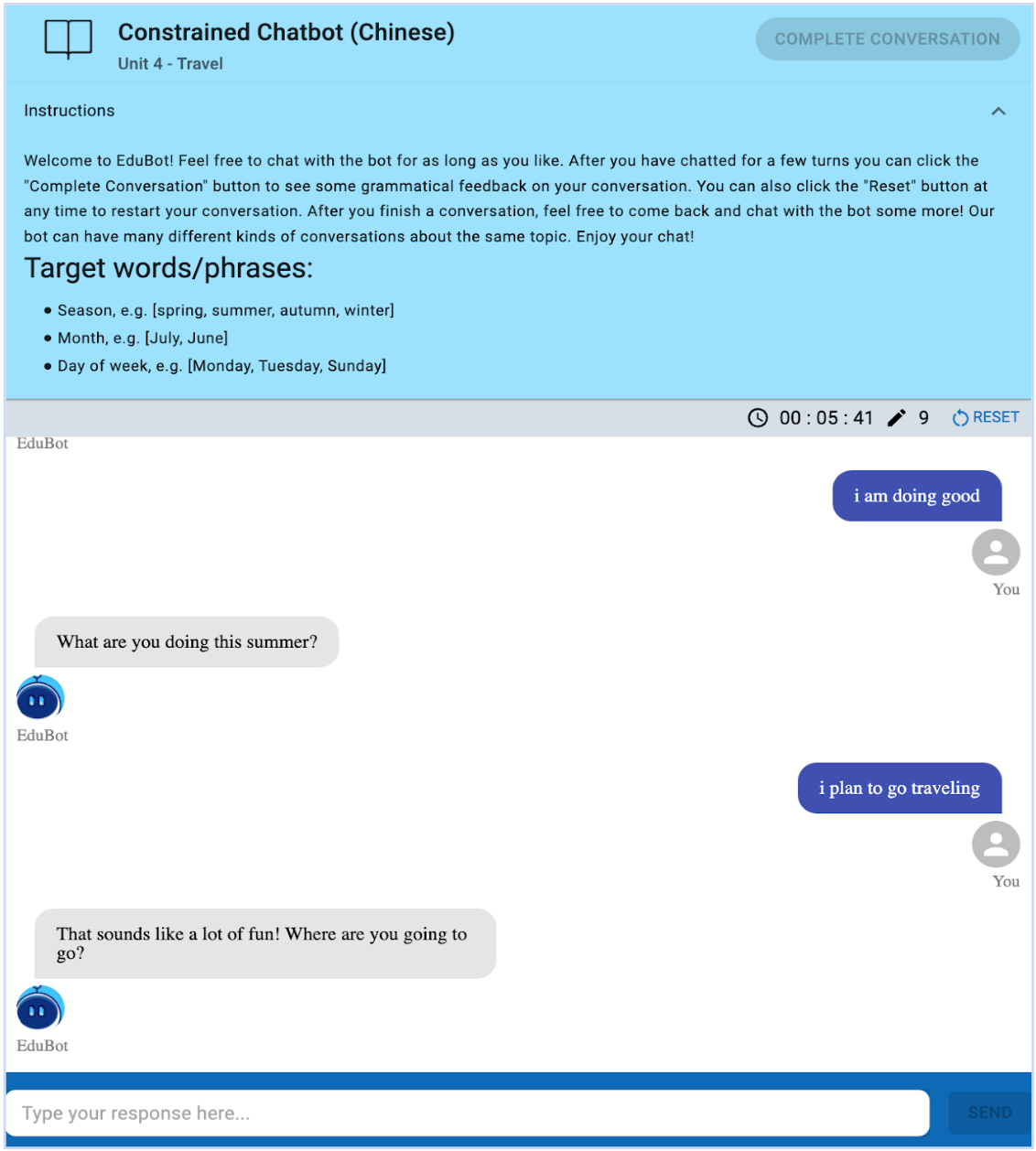}
         \caption{Conversation}
         \label{fig: conv}
     \end{subfigure}
     \hfill
     \begin{subfigure}[b]{0.45\textwidth}
         \centering
         \includegraphics[width=\textwidth]{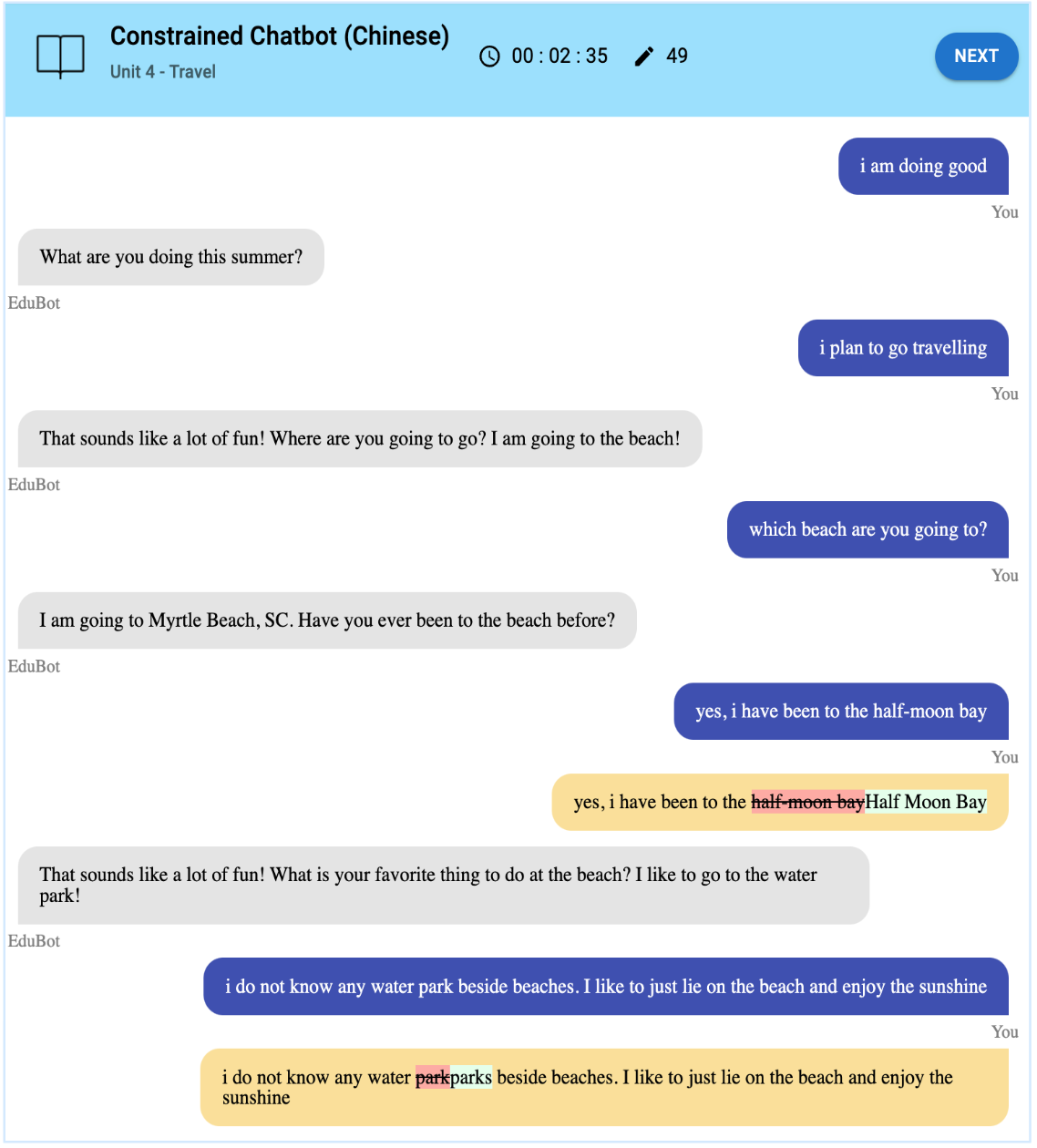}
         \caption{Grammar Error Correction}
         \label{fig: gec}
     \end{subfigure}
     \vfill
     \begin{subfigure}[b]{0.45\textwidth}
         \centering
         \includegraphics[width=\textwidth]{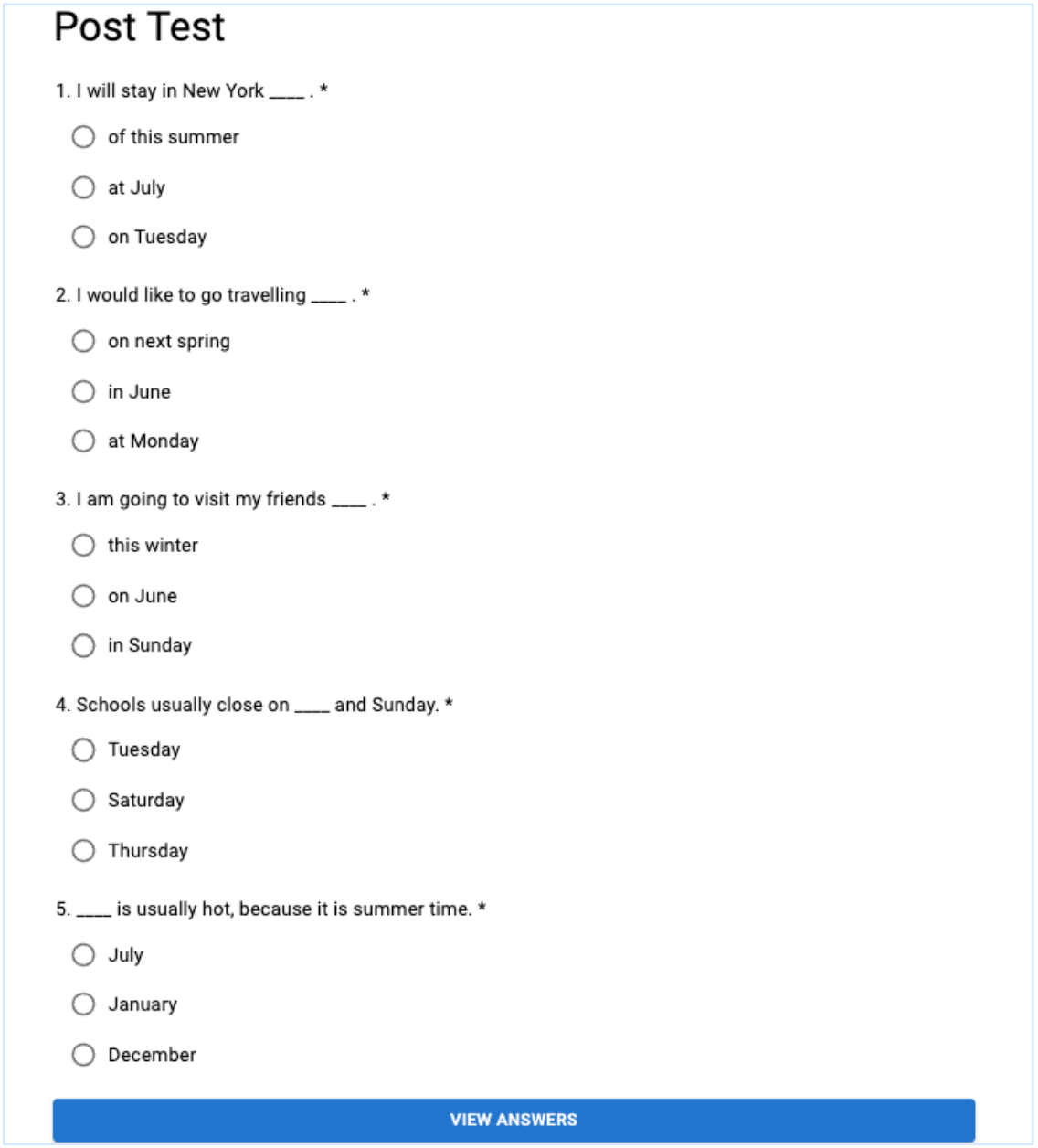}
         \caption{Post-Test}
         \label{fig: post-test}
     \end{subfigure}
     \hfill
     \begin{subfigure}[b]{0.45\textwidth}
         \centering
         \includegraphics[width=\textwidth]{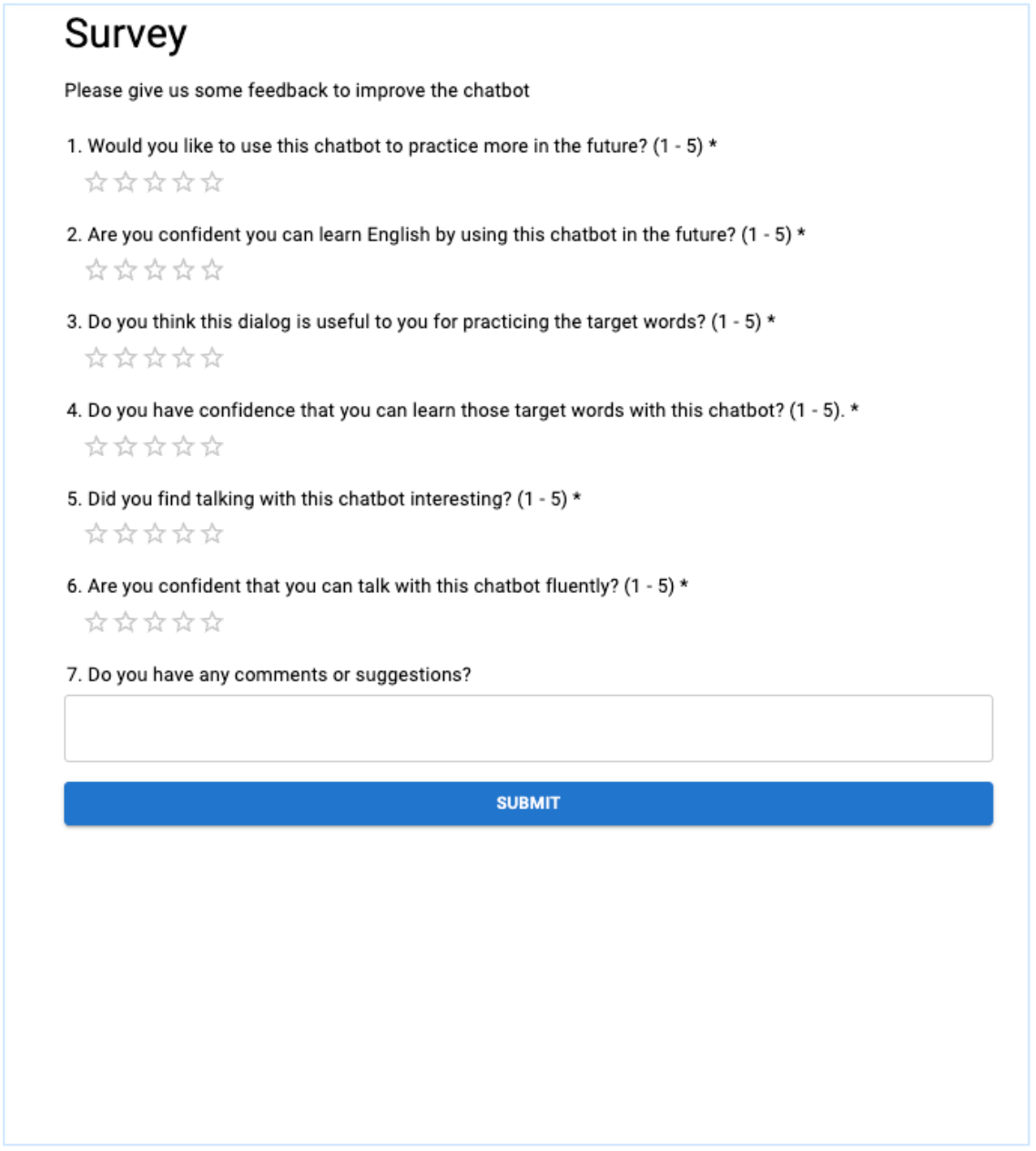}
         \caption{Survey}
         \label{fig: survey}
     \end{subfigure}
        \caption{The six-step evaluation flow, including instruction, pre-test, conversation, grammar error correction, post-test, and survey}
        \label{fig: flow}
\end{figure}
\noindent\textbf{Generalizable Constraints.} As previously mentioned, appending constraint words to generated sequences has no specific requirements. Therefore, any word or phrase can be included in the curriculum. Even an uncommon word can be accepted as the dynamic constraint threshold increases by turn number. Besides vocabulary, grammar can also be incorporated as constraints. For example, teachers can design a curriculum of past tense verbs for students to practice past tense. A curriculum of nouns in singular or plural form can help students distinguish between these two forms. Our framework can be adapted to support both grammar and vocabulary learning.
%
%
%
%
\subsection{Evaluation Flow}
To collect feedback and evaluate our model with real users, we employ the EduBot~\cite{Li2022UsingCT} platform for deployment. The whole evaluation flow, as shown below in Fig.~\ref{fig: flow} consists of six steps in total: 
\begin{enumerate}
    \item \textbf{Instruction}. As shown in Fig.~\ref{fig: instruction}, the instruction page briefly describes the function of our chatbot, the evaluation flow, and the purpose of the test. In addition, constraint words and phrases are also presented.
    \item \textbf{Pre-Test}. In order to evaluate whether our chatbot can help improve the user's understanding of constraint words, we design five single-choice questions based on the constraint words. Shown in Fig.~\ref{fig: pre-test}, these questions are designed to simulate the questions in students' exams.
    \item \textbf{Conversation}. During the Conversation, an abstracted instruction is shown to users by default to indicate that they should pay attention to the constraint words. Users have the option to collapse this instruction by clicking the ``$\string^$'' button. Conversation time and word counting are presented at the top of the conversation window. A ``reset'' button is also provided in case user wants to restart the conversation.
    \item \textbf{Grammar Error Correction}. Grammar error correction is a default function provided by the EduBot platform. It detects grammar errors from the user utterances and presents corrections at the end of conversations automatically. This function is technically supported by \cite{Yuan2021ErAConDEA}.
    \item \textbf{Post-Test}. As a comparison, we ask users to answer the same five questions as presented in the pre-test session. In order to demonstrate that our chatbot helps users learn constraint words/phrases, we expect users to correctly complete the post-test, even if they make mistakes during the pre-test session.
    \item \textbf{Survey}. Following \cite{Fryer2019ChatbotLP}, we design six survey questions for both self-efficacy and user interest. Users are asked to choose a score from one to five for each question and optionally leave comments or suggestions in a text box. 
\end{enumerate}
In Fig.~\ref{fig: flow}, all sentences are presented in English. However, during the evaluation, we translate the instructions (including the abstracted instructions during the conversation session) and survey sessions into Chinese in order to avoid misunderstandings and ensure the quality of the evaluation.
\section{Experiment and Results}
\subsection{Implementation Details}
Considering user volume and corresponding server capacity, we use the smallest version of BlenderBot 3, Blenderbot-3B, as our backbone model. 
Since constrained decoding consumes more time than normal decoding, while a deployable chatbot system requires low latency for more effective conversations, we set the beam size as ten. To encourage the model to generate long sequences, we adopt a length penalty of one. We also initialize the dynamic constraint threshold as ten so our model can generate more flexible utterances at the beginning of the conversation.

From our consultations with students' English teachers, we chose words of seasons (e.g. spring, winter), months (e.g. July, December), and days (e.g. Monday, Sunday) as constraint words. To enforce the model to use these words with their matching preposition, we also include constraint phrases such as ``on Monday'' or ``during winter''. We adopt a disjunctive format for constraint words, meaning that only one word from each type (seasons, months, or days) is generated. Once a constraint word is generated, the whole type would be marked as satisfied and removed from the constraint list~(Sec.\ref{sec:grid}).
\subsection{Results and Analysis}
\begin{figure}[h]
\centering
\includegraphics[width=0.6\textwidth]{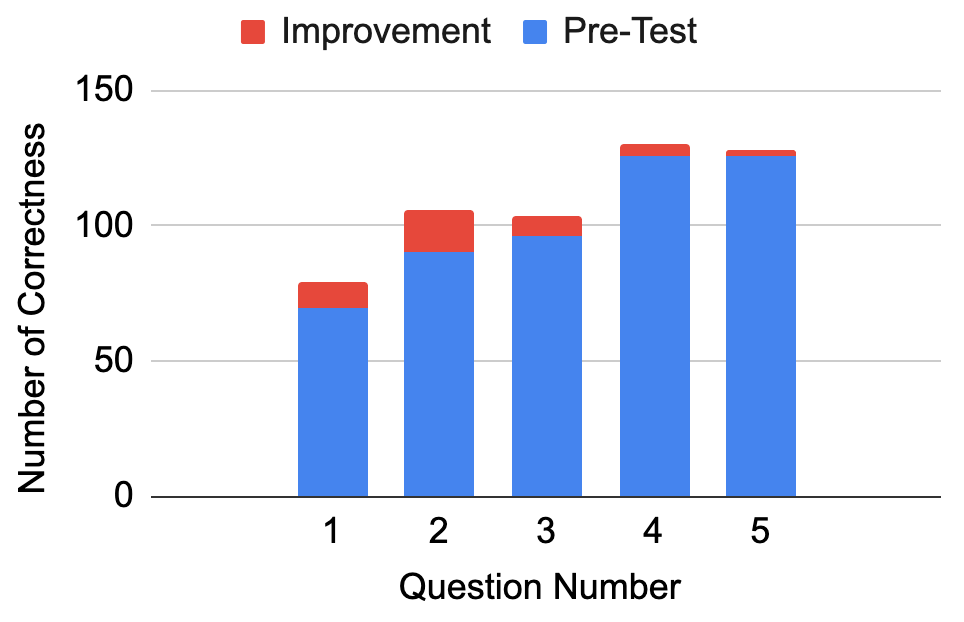}
\caption{The correctness of each test question from 155 8th-grade Chinese students. The blue bar represents the number of students that correctly answered questions in the pre-test. The red bar indicates the number of students who failed the corresponding question in the pre-test but correctly answered it after the conversation.} \label{fig: test_result}
\vspace{-0.3cm}
\end{figure}

\noindent\textbf{Conversation helps answer test questions.} In order to evaluate students' understanding of constraint words in terms of both grammar and vocabulary, we designed the first three pre-test questions for preposition usage of corresponding constraint words. While the last two questions are designed to test whether students understand the semantic meaning of constraint words. We recruit 155 8th-grade Chinese students for evaluation in total. The results in Fig.~\ref{fig: test_result} show that around half of students can correctly answer the first three questions and 80\% of students give correct answers for the final two questions. This suggests that most of the students understand the semantic meaning of constraint words but aren't able to use them with correct propositions.
%
%
The red bar in Fig.~\ref{fig: test_result} represents how many students made a mistake on the given question in the pre-test but gave the correct answer after their conversation. In total, 39 out of 267 (14.61\%) incorrect answers were corrected after speaking with our chatbot. 
We also count that 21 out of 155 students ($13.5\%$) improve their overall score for all five test questions.
This improvement demonstrates that our chatbot help students learn both semantic meaning and usage of constraint words and phrases.

%
%
\noindent\textbf{Constrained decoding encourages users to practice target words more frequently.} We also count the usage frequencies of constraint words in both user utterances and system responses. The result is listed in Table~\ref{tab: freq}. Since we adopt a dynamic constraint threshold (Sec.~\ref{sec:grid}) during decoding, which does not force our model to use all constraints at the beginning of a conversation. On the other hand, not all students complete a conversation after seeing all constraint words, which means not all constraint words are necessarily used during the course of a conversation. This leads to the system's usage frequency of constraint words being less than 100\%. However, it is obvious that both the system and user sides use words for ``season'' more frequently than the other two types. This suggests that forcing system to generate constraint words encourages users to use these words in their own utterances. 
We also correlate the usage of constraint words and the improvement of testing results. 
Specifically, we compute the frequency of constraint word usage over those students who made wrong answers in the pre-test but correctly answered them after the conversation (same as the students of the red bar in Fig.~\ref{fig: test_result}).
We find that both the system and users are more likely to use constraint words in the improved test cases. In other words, more frequently seeing and practicing constraint words help users learn those words.

\begin{table}[t]
\centering
\normalsize
\caption{Frequencies of constraint word usage for both system side and user side. Percentages on the left side are counted over all 155 test cases. Percentages on the right half are counted over dialogs after which students corrected their test answers. ``Seasons'', ``Months'' and ``Days'' are the three constraint word types. ``Any'' means any type of constraint words mentioned in utterances counts.}
\begin{tabular}{l|cccc|cccc}
\toprule
       & \multicolumn{4}{c|}{All Test Cases}    & \multicolumn{4}{c}{Improved Test Cases} \\
       \midrule
Constraint & Seasons & Months  & Days    & Any     & Seasons  & Months  & Days    & Any      \\
       \midrule
System & 81.94\% & 65.81\% & 65.16\% & 90.32\% & 90.00\%  & 73.33\% & 66.67\% & 100.00\% \\
User   & 38.71\% & 18.71\% & 24.52\% & 50.32\% & 40.00\%  & 30.00\% & 26.67\% & 96.67\%  \\
\bottomrule
\end{tabular}
\label{tab: freq}
\vspace{-0.3cm}
\end{table}

\begin{table}[h]
\vspace{-0.5cm}
\centering
\scriptsize
\caption{The average score of each survey question.}
\begin{tabular}{l|c}
\toprule
Questions                                                          & Scores \\
\midrule
1. Would you like to use this chatbot to practice more in the future? & 4.34                      \\
2. Are you confident you can learn English  by using this chatbot in the future?        & 4.36                      \\
3. Do you think this dialog is useful to you for practicing the target words?    & 4.44                      \\
4. Do you have confidence that you can learn those target words with this chatbot?     & 4.38                      \\
5. Do you find talking with this chatbot interesting?  & 4.36                      \\
6. Are you confident that you can talk with this chatbot fluently?  & 4.39    \\ 
\bottomrule
\end{tabular}
\label{tab: survey}
\vspace{-0.3cm}
\end{table}
\begin{figure}[h]
\vspace{-0.2cm}
\centering
\includegraphics[width=0.6\textwidth]{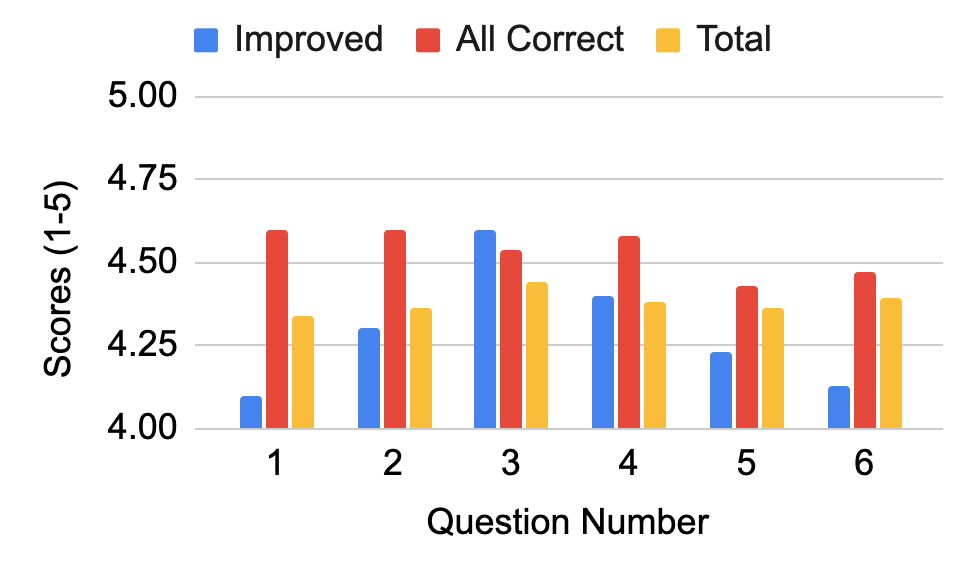}
\caption{Average scores for six survey questions. The blue bar (``Improved'') computes over those who make mistakes in the pre-test but correct their answer in the post-test, and the red bar (``All Correct'') represents the scores of those who correctly answer all questions in the pre-test. The orange one (``Total'') computes scores over all students } \label{fig: survey_result}
\vspace{-0.5cm}
\end{figure}
\noindent\textbf{Students recognize the functionality of the chatbot.} Table~\ref{tab: survey} lists all six survey questions and their average scores. In general, students recognize that our chatbot helps them learn English. To dig deeper, we further compute scores for those who corrected their answers after conversations and those who correctly answered all questions in the pre-test (repesented by blue and red bar correspondingly in Fig.~\ref{fig: survey_result}). The scores of red bar are above the total average scores (orange bar) over all six questions. This suggests that students who can easily solve test questions hold positive altitudes towards our chatbot and enjoy the conversation with it. On the other hand, the scores represented by the blue bar only surpass the average for questions three and four, which are focused on the learning effect. This indicates that those students are less confident to talk with our chatbot. But they do acknowledge that it is helpful in learning the constraint words. Overall, this once again confirms that incorporating a curriculum through constrained decoding helps students learn specified words.






\section{Conclusion}
In this paper, we propose a user-adaptive generative chatbot for language learning. We use constrained decoding to incorporate a curriculum, which is adaptable based on the user's request. We apply this method to a pre-trained large language model. To evaluate our model, we design an evaluation flow based on constraint words and employ more than 155 students who learn English as a second language. The result demonstrates that our curriculum-incorporated chatbot help students learn specific words and phrases.

%
%
%
\bibliographystyle{splncs04}
\bibliography{custom}

\end{document}